\documentclass[runningheads]{llncs}
\usepackage{xcolor}
\usepackage{amsmath}
\usepackage{graphicx}
\usepackage{amssymb}
\usepackage{multirow}
\usepackage{caption}
\usepackage{subcaption}
\usepackage{array}
\usepackage[pagebackref=true,breaklinks=true,colorlinks,bookmarks=false]{hyperref}

\usepackage[T1]{fontenc}
\def\doi#1{\href{https://doi.org/\detokenize{#1}}{\url{https://doi.org/\detokenize{#1}}}}
\usepackage{graphicx}

\newcolumntype{!}{>{\global\let\currentrowstyle\relax}}

\newcolumntype{^}{>{\currentrowstyle}}
\newcommand{\rowstyle}[1]{\gdef\currentrowstyle{#1}%

	#1\ignorespaces

}

\begin{document}
\title{Improving Localization for Semi-Supervised Object Detection}
%
%
\author{Leonardo Rossi\orcidID{0000-0002-9316-595X} \and
        Akbar Karimi\orcidID{0000-0002-5132-2435} \and
        Andrea Prati\orcidID{0000-0002-1211-529X}}
%
%
\institute{IMP Lab - D.I.A. - University of Parma - Parma, Italy\\
\email{\{leonardo.rossi, akbar.karimi, andrea.prati\}@unipr.it}\\
\url{implab.ce.unipr.it/}}
\maketitle              
\begin{abstract}
Nowadays, Semi-Supervised Object Detection (SSOD) is a hot topic, since, while it is rather easy to collect images for creating a new dataset, labeling them is still an expensive and time-consuming task.
One of the successful methods to take advantage of raw images on a Semi-Supervised Learning (SSL) setting is the Mean Teacher technique \cite{Tarvainen2017MeanTA}, where the operations of pseudo-labeling by the Teacher and the Knowledge Transfer from the Student to the Teacher take place simultaneously.
However, the pseudo-labeling by thresholding is not the best solution since the confidence value is not strictly related to the prediction uncertainty, not permitting to safely filter predictions.
In this paper, we introduce an additional classification task for bounding box localization to improve the filtering of the predicted bounding boxes and obtain higher quality on Student training.
Furthermore, we empirically prove that bounding box regression on the unsupervised part can equally contribute to the training as much as category classification.
Our experiments show that our IL-net (Improving Localization net) increases SSOD performance by 1.14\% AP on COCO dataset in limited-annotation regime.
The code is available at \url{https://github.com/IMPLabUniPr/unbiased-teacher/tree/ilnet}.

\end{abstract}
\section{Introduction}
\label{introduction}
Supervised learning usually requires a large amount of annotated training data which can be expensive and time-consuming to produce.
Semi-Supervised Learning (SSL), on the other hand, addresses this issue by taking advantage of large unlabeled data accompanied by a small labeled dataset.
Usually, in an SSL setting, we have two models that act as Teacher and Student.
The latter is trained in a supervised way, drawing the needed ground truths from the dataset, if exist, and from the Teacher's predictions, otherwise.

This approach can be beneficial in many machine learning tasks including object detection, whose final result is a list of bounding boxes (bboxes) and the corresponding classes.
In this task, the network is devoted to finding the position of the localized objects in an image, as well as identifying which class they belong to.
Hereinafter, we will refer to this specific application as Semi-Supervised Object Detection (SSOD). In \cite{liu2021unbiased}, authors have collected many interesting ideas applied to SSOD.
Among them, the one that seems more promising is the Mean Teacher \cite{Tarvainen2017MeanTA}, which uses the Exponential Moving Average (EMA) as a knowledge transfer technique from the Student to the Teacher.
In particular, while the Teacher produces pseudo-labels on unlabeled and weakly augmented images to obtain more reliable labels, the Student is trained using the pseudo-labels as ground-truth on the same images, but strongly augmented to differentiate the training.
In weak augmentation, only a random horizontal flip is applied, while in strong augmentation, other transformations such as randomly adding color jittering, grayscale, Gaussian blur, and cutout patches are also used.

Although they obtain good performances with this pseudo-labeling technique, the Student model is still influenced by the Teacher's erroneous predictions.
The quality of these predictions is difficult to control by applying only a simple confidence thresholding.
For this reason, there is a need for an additional way to strengthen the region proposals and reduce the number of erroneous predictions by the Teacher.
In the proposed architecture, we introduce a new task used for the classification of the bboxes, with the aim of distinguishing good quality ones from the others.
This new score exploits the information complementary to the class score already used in these networks, allowing a different level of filtering.  In addition, we show how to take advantage of the regression tasks on the unsupervised learning part.
Usually, they are excluded in the unsupervised training phase.
The justification is that the classification score is not able to filter the potentially incorrect bboxes \cite{jiang2018acquisition,liu2021unbiased}.
A well-known problem during training is the Objective Imbalance \cite{oksuz2020imbalance}, which is characterized by the difficulty in balancing the different loss contributions.
In our hypothesis, this is the case for the unsupervised training part.
In order to obtain a positive effect from regression losses, an adequate balance of the contribution of these two losses (regression and category classification) is a possible solution to the above-mentioned problem.
In this way, we prevent the regression losses on unsupervised dataset from dominating the training, a phenomenon that greatly amplifies the error introduced by inaccurate Teacher predictions.
\\

The main contributions of this paper are the following:
\begin{itemize}
	\item a new bounding box IoU (Intersection over Union) classification task to filter out errors on pseudo-labels produced by the Teacher;
	\item the introduction of a regression task on the unlabeled dataset which can help the network to learn better;
	\item an exhaustive ablation study on all the components of the architecture.
\end{itemize}

\section{Related Work}
\label{related}

\noindent\textbf{Semi-Supervised Learning for Object Detection}.
In \cite{samuli2017temporal}, the authors proposed the $\Pi$-Model which is an ensemble of the predictions of the model at different epochs under multiple regularizations and input augmentation conditions. The predictions for the unlabeled images are merged together to form a better predictor.
In \cite{jeong2019consistency}, the authors proposed Consistency-based Semi-supervised Learning for Object Detection (CSD), which uses consistency constraints as a self-training task to obtain a better training of the network.
In \cite{sohn2020simple}, an SSL framework with Self-Training (via pseudo label) and the Augmentation driven Consistency regularization (STAC) is introduced, exploiting weak data augmentation for model training and strong data augmentation for pseudo-labeling.
Several works \cite{liu2021unbiased,tang2021humble,Tarvainen2017MeanTA} proposed to use the Mean Teacher model, applying EMA to the weights instead of the predictions, facilitating knowledge transfer from Student to the Teacher model at each iteration.
For one-stage object detection models, authors of \cite{nguyen2019semi} use an Expectation-Maximization approach, generating pseudo-labels in the expectation step and training the model on them in the maximization step, optimizing for classification in each iteration and for localization in each epoch.
In \cite{xu2021end}, the authors propose a Soft Teacher mechanism where the classification loss of each unlabeled box is weighted by the Teacher classification score, also using a box jittering approach to select the most reliable pseudo-bboxes.
In \cite{li2020improving}, authors utilize SelectiveNet to properly filter pseudo-bboxes trained after the Teacher.
In \cite{zhou2021instant}, the authors propose two models, one of which generates a proposal list of bounding boxes and the second one refines these proposals, using the average class probability and the weighted average of the bounding box coordinates.

\noindent\textbf{Bounding Box Intersection over Union (BBox IoU)}.
In \cite{he2019bounding}, the authors added a new branch on top of the Fast R-CNN model to estimate standard deviation of each bounding box and use it at Non-maximum Suppression (NMS) level to give more weight to the less uncertain ones.
In \cite{jiang2018acquisition,zhu2021iou}, the authors added a new branch on top of the Faster R-CNN to regress bounding box IoU and to multiply this value with the classification score to compute final score of the suppression criterion of the NMS.
In the same direction, Fitness NMS was proposed in \cite{tychsen2018improving} to correct the detection score for better selecting bounding boxes which maximize their estimated IoU with the ground-truth.
\section{Teacher-Student Learning Architecture}
\label{ts-learning-arch}
In SSOD, we have two datasets.
The first set $\mathbf{D_s} = \{x_i^s, y_i^s\}_{i=1}^{N_s}$, typically smaller, contains $N_s$ images $\mathbf{x^s}$ with the corresponding labels $\mathbf{y^s}$, and the second set $\mathbf{D_u} = \{x_i^u\}_{i=1}^{N_u}$ contains $N_u$ images $\mathbf{x^u}$ without labels. Similarly to \cite{liu2021unbiased}, our architecture is composed of two identical models where one behaves as Teacher and the other as the Student.
At each iteration of the semi-supervised training, the Student is trained with one batch of images coming from $\mathbf{D_s}$ (supervised training) and one from $\mathbf{D_u}$ (unsupervised training), using the pseudo-labels generated by the Teacher as ground-truth and filtered by a threshold.
Then, the Student transfers the learned knowledge to the Teacher using the EMA applied to the weights.

In our model, the total loss $L$ for the Student is composed of two terms, which come from the the supervised $L_{sup}$ and unsupervised $L_{unsup}$ training:

\begin{equation}
\begin{split}
L &= L_{sup} + L_{unsup} \\
L_{sup} &= \sum_{i} L_{cls} + L_{reg} + \boldsymbol{L^{IoU}}\\
L_{unsup} &= \sum_{i} \alpha L_{cls} + \boldsymbol{\beta L_{reg}} + \boldsymbol{\gamma L^{IoU}} \\
L_{cls} &= L_{cls}^{rpn}(x_i^s, y_i^s) + L_{cls}^{roi}(x_i^s, y_i^s) \\
L_{reg} &= L_{reg}^{rpn}(x_i^s, y_i^s) + L_{reg}^{roi}(x_i^s, y_i^s) \\
\end{split}
\label{eq:all_losses}
\end{equation}

\noindent where $L_{cls}$ contains the sum of Region Proposal Network (RPN) and Region of Interest (RoI) classification losses, while $L_{reg}$ contains the sum of RPN and RoI regression losses.
The $L^{IoU}$ loss will be defined in Subsection \ref{sec:bbox_iou}.
$\alpha$, $\beta$ and $\gamma$ represent how much weight each component of the unsupervised training has.
The new terms introduced in this paper, w.r.t. the ones defined in \cite{liu2021unbiased}, are shown in bold in the above equations and will be described in the following.

\subsection{Bounding Box Regression on Unsupervised Training}
The training consists of two phases, the \textit{burn up} stage and \textit{Teacher-Student Mutual Learning} stage.
In the first stage, the Student is trained only on the labeled dataset, following the standard supervised training procedure.
Then, for the second stage, at each iteration, the Student is trained on two batches at the same time, one coming from the labeled dataset and the other coming from the unlabeled dataset.
For the latter, the baseline \cite{liu2021unbiased} trains only the RPN and RoI head classifier and disregards regression.
The authors justify this choice noticing that the classification confidence thresholds are not able to filter the incorrect bounding box regression.
Our hypothesis is that training also on pseudo bounding box regression could help the Student as long as the task is correctly weighted with respect to the others.
Category classification and regression tasks are learned in parallel.
Given that the classification performance improves during the training, we can also expect the bounding box regression to behave the same way.
In other words, while the training proceeds, the average quality of the pseudo-labels increases and so does the classification confidence value.

\begin{figure*}[h!]
	\centering
	\begin{subfigure}{0.43\linewidth}
		\includegraphics[width=1.\linewidth]{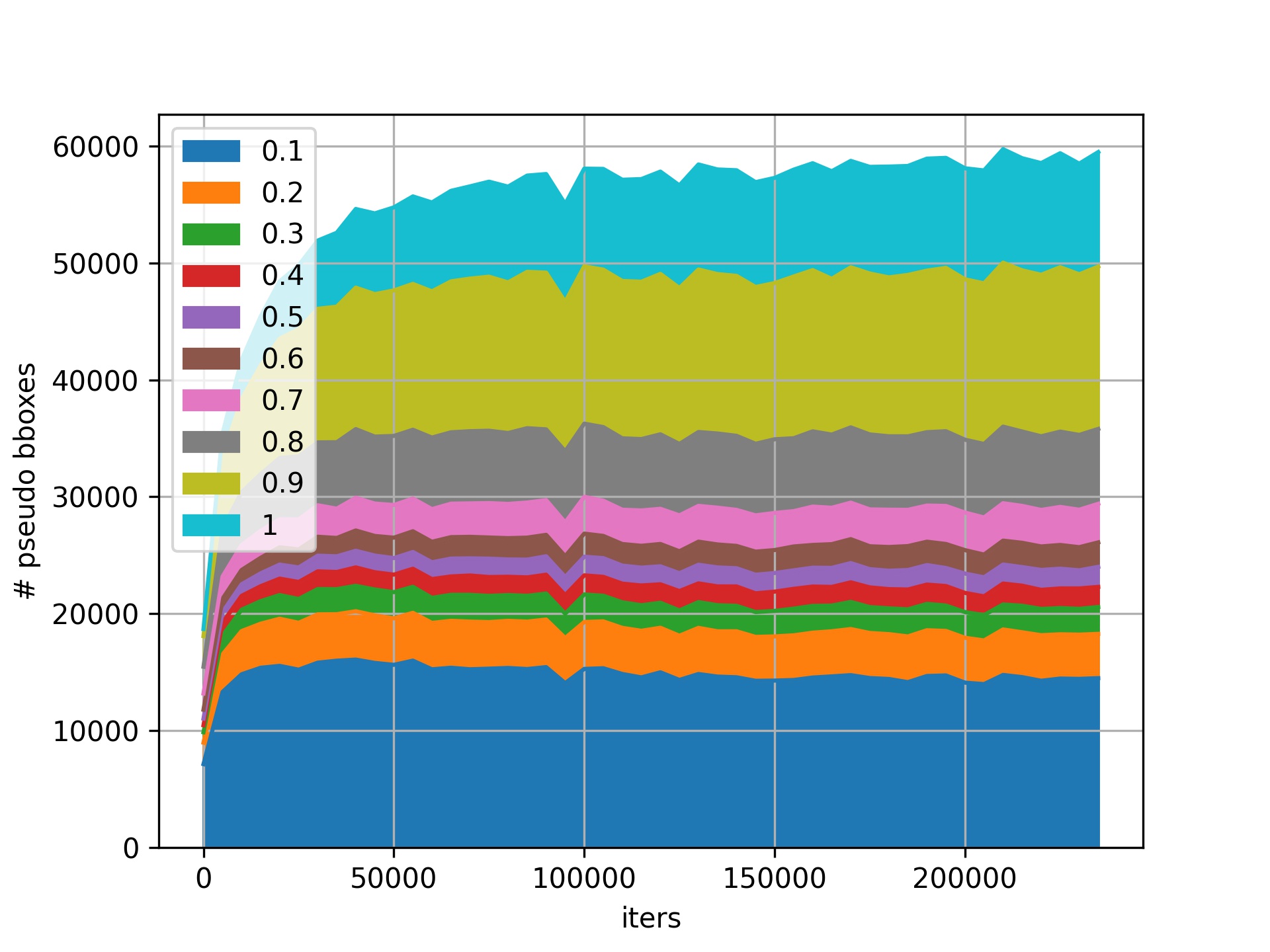}
		\caption{count pseudo bboxes}
		\label{fig:psuedo-bboxes}
	\end{subfigure}
	\begin{subfigure}{0.43\linewidth}
		\includegraphics[width=1.\linewidth]{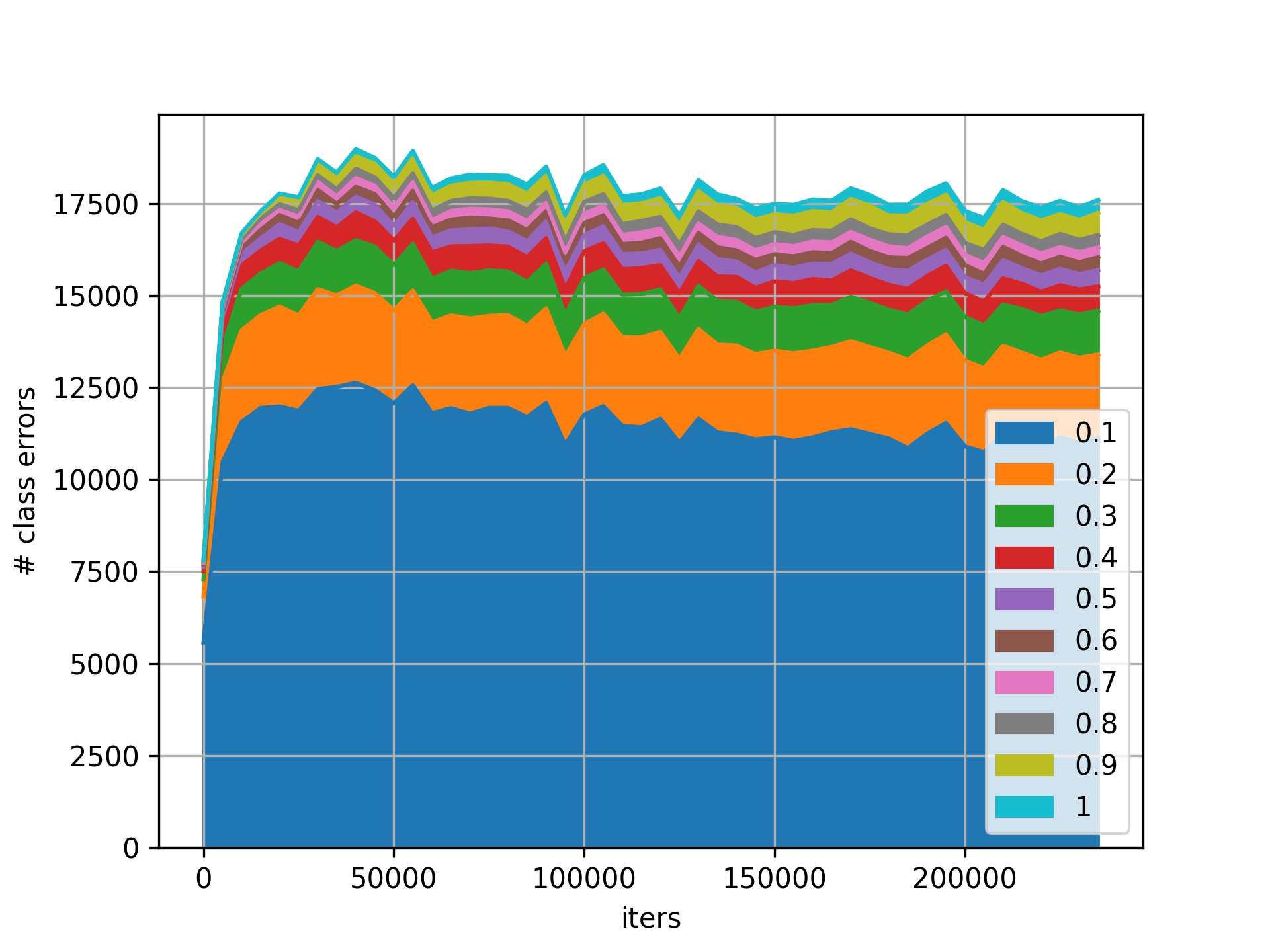}
		\caption{count class errors}
		\label{fig:class-errors}
	\end{subfigure}
	\begin{subfigure}{0.43\linewidth}
		\includegraphics[width=1.\linewidth]{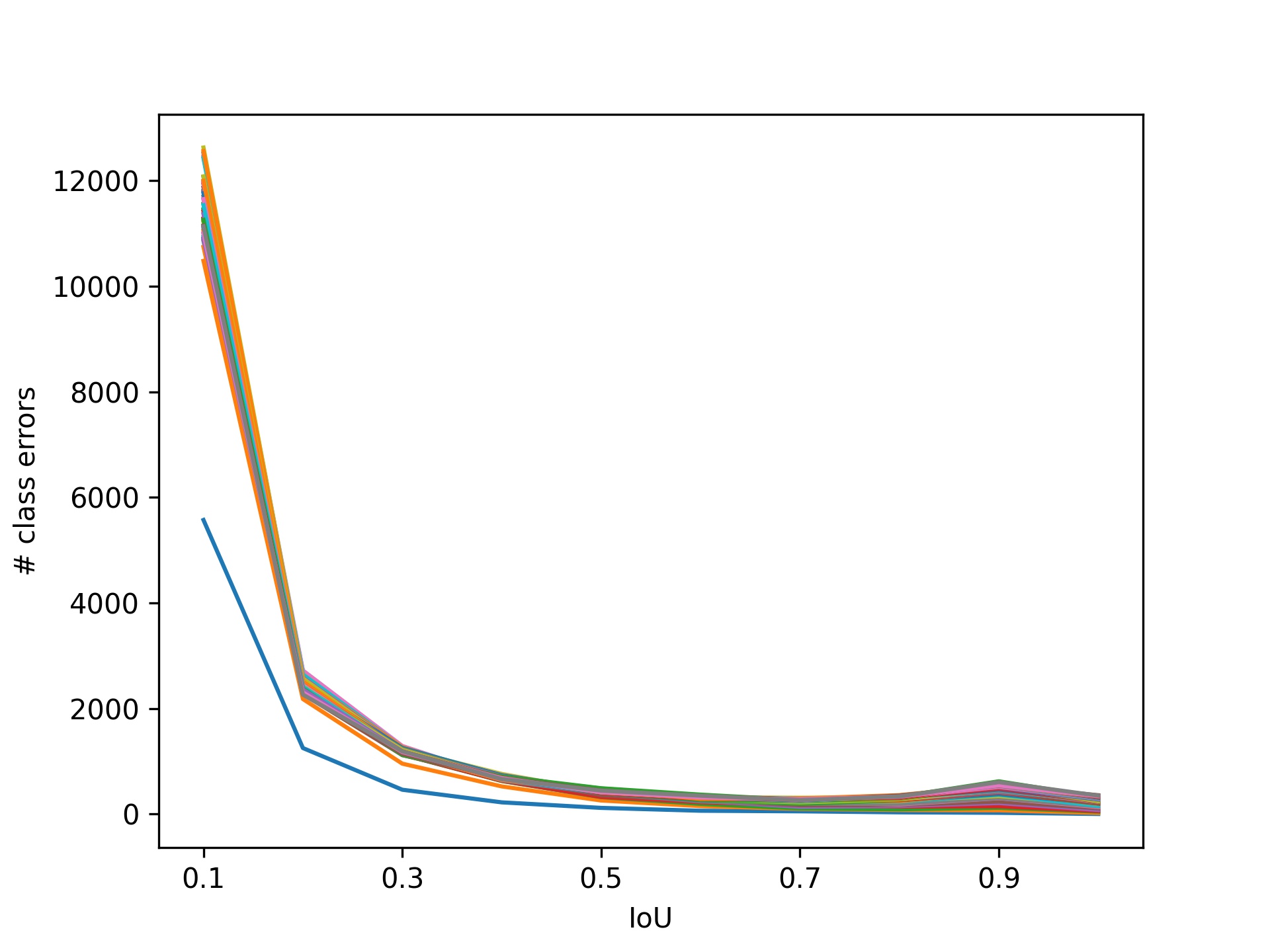}
		\caption{count class errors per IoU}
		\label{fig:class-errors-exp}
	\end{subfigure}
	\begin{subfigure}{0.43\linewidth}
		\includegraphics[width=1.\linewidth]{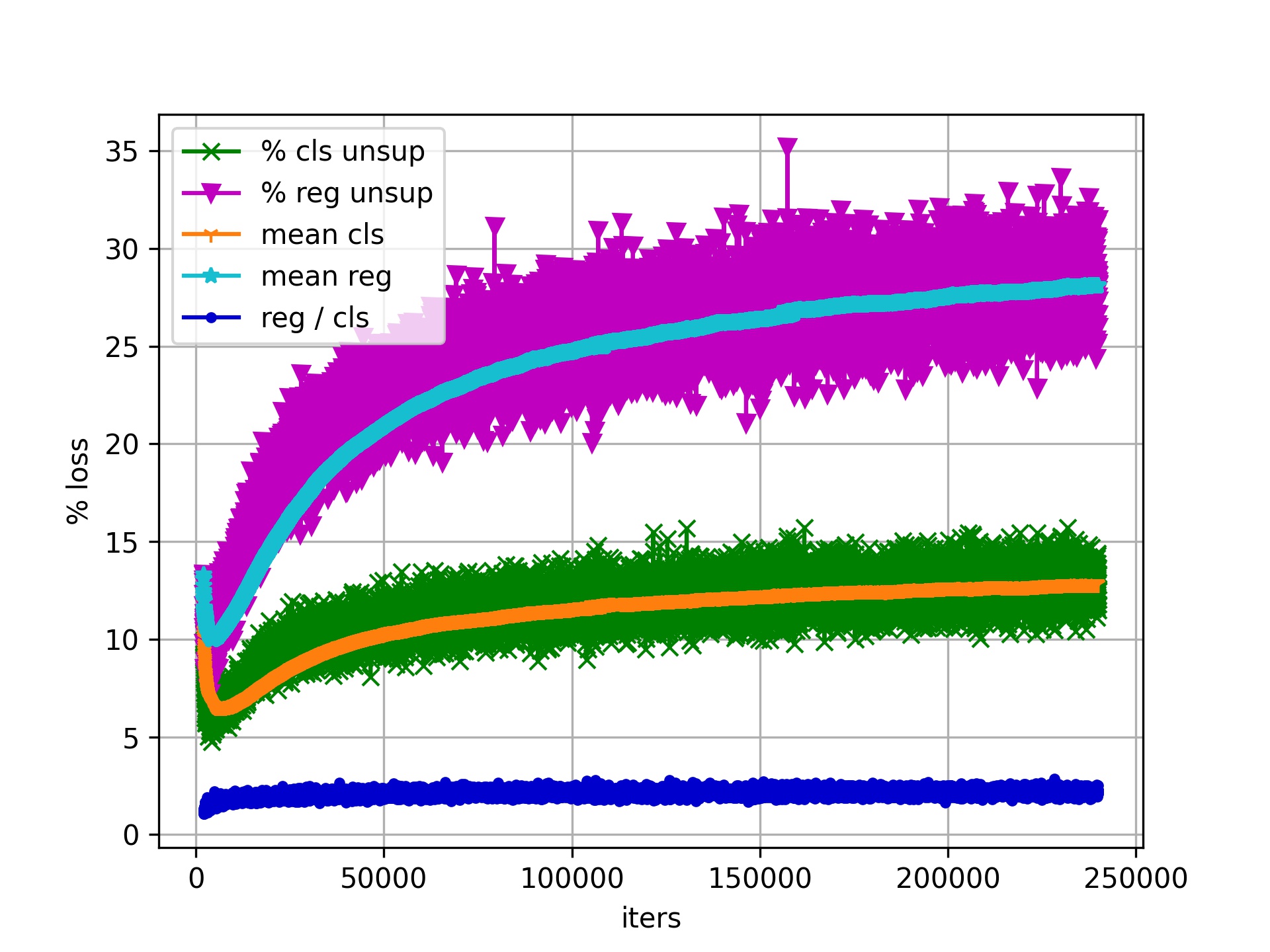}
		\caption{Unsupervised $L_{cls}$ and $L_{reg}$}
		\label{fig:compare-cls-reg-losses}
	\end{subfigure}
	\caption{The pseudo bounding boxes generated during training: (\ref{fig:psuedo-bboxes}) IoU distribution of pseudo-bboxes, (\ref{fig:class-errors}) distribution of pseudo-bboxes when the predicted class is wrong. (\ref{fig:class-errors-exp}) number of pseudo-bboxes per IoU collected every 5000 iterations.
	(\ref{fig:compare-cls-reg-losses}) Unsupervised classification and regression losses comparison during the training.
		Better seen in color.}
	\label{fig:iou-bboxes}
\end{figure*}

During the Student training, although pseudo bounding boxes are filtered with a threshold on classification confidence score, we still have bounding boxes of any IoU quality.
This problem is related to the uncertainty on prediction.
Figure \ref{fig:psuedo-bboxes} visualizes the IoU distribution quality with respect to the ground-truth of the Teacher filtered predictions.
We can notice that the number of pseudo bounding boxes with IoU less than the threshold (0.6 in our experiments) remains almost constant during the entire training, unlike the others which slowly increase.
In Figure \ref{fig:class-errors}, we show only the filtered pseudo bounding boxes that the Teacher has wrongly classified and split by their IoU.
In Figure \ref{fig:class-errors-exp}, the same data are shown using a different graph.
We can notice that the number of wrongly classified pseudo-bboxes decreases exponentially with the increase of the quality, and this trend remains the same during the training.
The bounding boxes with low quality ($IoU < 0.6$) represent 45\% of the total pseudo-bboxes and more than 90\% of classification errors.
This means that the low-quality IoU bboxes contain almost all the classification errors.

By looking at Figure \ref{fig:compare-cls-reg-losses}, we can see that the unsupervised regression loss (from RPN and RoI heads) represents between 20\% and 30\% of the total loss.
Conversely, the unsupervised classification loss (from RPN and RoI heads) accounts only for 15\% (at most) of the total loss.
This means that an error in pseudo-labels has almost three times (see blue line in Figure \ref{fig:compare-cls-reg-losses}) more weight on regression branch than on classification branch.
To avoid the amplification of the errors, an appropriate value for $\beta$ is chosen, by under-weighting the regression contribution and making it comparable with the classification one.

\subsection{Bounding Box IoU Classification Task}
\label{sec:bbox_iou}

As noted by \cite{jiang2018acquisition,zhu2021iou}, adding a new task related to the IoU prediction could help the network to train better.
Our model learns to predict the bounding box quality, using this value to further filter pseudo-labels in conjunction with the classification score.
Differently from them, we train the model to make a binary classification instead of a regression since it is sufficient (and easier to learn) for the purpose of filtering.

\begin{figure*}[t]
\centering
	\includegraphics[trim=40 580 0 75, clip, width=1.5\linewidth]{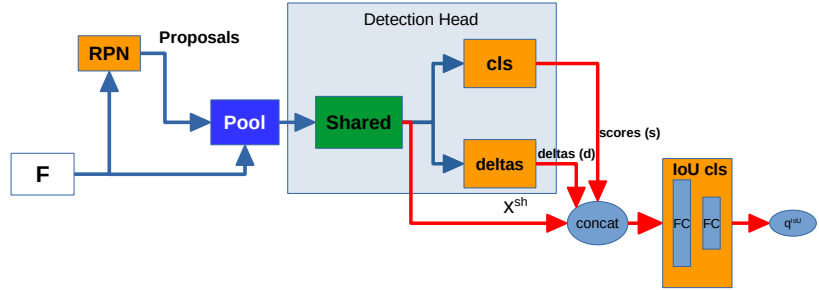}
	\caption{Faster R-CNN architecture with our branch in red.}
	\label{fig:bbox-iou-arch}
\end{figure*}

Fig. \ref{fig:bbox-iou-arch} illustrates the architecture of a two-stage object detection model such as Faster R-CNN \cite{ren2015faster} with our additional branch highlighted in red. For each positive bounding box (\textit{i.e.}, the ones recognized as foreground), it concatenates the output of the shared layers ($x_i^{sh}$, with the size of 1024), all the classification scores ($s_i$, with the size of the number of classes plus the background) and all bounding box regression deltas ($d_i$, with the size of 4).
All these features are passed to the \textit{IoU cls} branch (also called $f(\cdot)$ in the following), which outputs a vector with the same size as $s_i$ (\textit{i.e.}, one for each class). This will return the $i^{th}$ IoU classification, called \textit{IoU score} $q^{IoU}$, corresponding to the predicted bbox.
The $f(\cdot)$ branch consists of two fully-connected layers with an ELU \cite{clevert2015fast} activation function between them and a \textit{sigmoid} activation function at the end.

The loss $L_i^{IoU}$ for the new branch of IoU classification, conditioned on the class, is defined as:

\begin{equation}
\begin{split}
	q_{i}^{IoU} &= f(concat(x_i^{sh}, s_i, d_i)) \\
	L_i^{IoU} &= FL(q_{i}^{IoU},\: t_i) \\
\end{split}
\label{eq:iou_branch}
\end{equation}

\noindent where the FL function is the Focal Loss \cite{lin2017focal} with its own $\gamma$ value equal to 1.5 and $t_i$ represents the binary target label.
For the $i^{th}$ bounding box, the branch output $q_{i}^{IoU} \in [0, 1]$ is a single value which predicts if it is a high- or low-quality bounding box. The target $t_i$ and its IoU are defined as follows:

\begin{equation}
\begin{split}
    IoU_i &= \max{\textrm{IoU}\left(b_i^t,g\right)}\:\:\:\:\:\forall b_{i} \in \mathbf{B} | IoU\left(b_i,g\right) \geqslant u \\
	t_i &=
	\begin{cases}
		1  & IoU_i > \mu \\
		0  & \textrm{otherwise} \\
	\end{cases}
\end{split}
\end{equation}

\noindent where $\mathbf{B}$ is the list of proposals coming from the RPN, $b_i$ are the $i^{th}$ predicted bounding boxes, $g$ is the ground-truth bounding box, $u$ is the minimum IoU threshold to consider the bounding $b_i$ as positive example (typically set to 0.5) and $\mu$ (set to 0.75) is the minimum IoU threshold to be classified as high-quality.

In the evaluation phase, the Teacher's filtering of the predicted bounding boxes with low confidence is preceded by the IoU classification filtering, which uses a new IoU inference threshold $\theta$ (set to 0.4).



%

\section{Experiments}
\label{experiments}

\noindent\textbf{Dataset}.
We perform our tests on the MS COCO 2017 dataset \cite{lin2014microsoft}.
The training dataset consists of more than 117,000 images and 80 different classes of objects, where 10\% of the labeled images are used.

\noindent\textbf{Evaluation Metrics}.
All the tests are done on COCO minival 2017 validation dataset, which contains 5000 images.
We report mean Average Precision (mAP) and $AP_{50}$ and $AP_{75}$ with 0.5 and 0.75 minimum IoU thresholds, and $AP_s$, $AP_m$ and $AP_l$ for small, medium and large objects, respectively.

\noindent\textbf{Implementation details.}
All the values are obtained running the training with the same hardware and hyper-parameters.
When available, the original code released by the authors is used.
Our code is developed on top of the Unbiased Teacher \cite{liu2021unbiased} source code.
We perform the training on a single machine with 6 Tesla P100 GPUs with 12GB of memory.
The train lasts 180,000 iterations with a batch size of 2 images per GPU for the supervised part and 2 images for the unsupervised part, with $\alpha$ set to 4.
We use the Stochastic Gradient Descent (SGD) optimization algorithm with a learning rate of 0.0075, a weight decay of 0.0001, and a momentum of 0.9.
The learning rate decays at iteration 179990 and 179995.
We use the Faster R-CNN with FPN \cite{lin2017feature} and the ResNet 50 \cite{he2016deep} backbone for the teacher and student models, initialized by pre-trained on ImageNet, and the same augmentation of Unbiased Teacher \cite{liu2021unbiased}.

\begin{figure*}[bt]
	\centering
	\begin{subfigure}{0.43\linewidth}
		\includegraphics[width=0.9\linewidth]{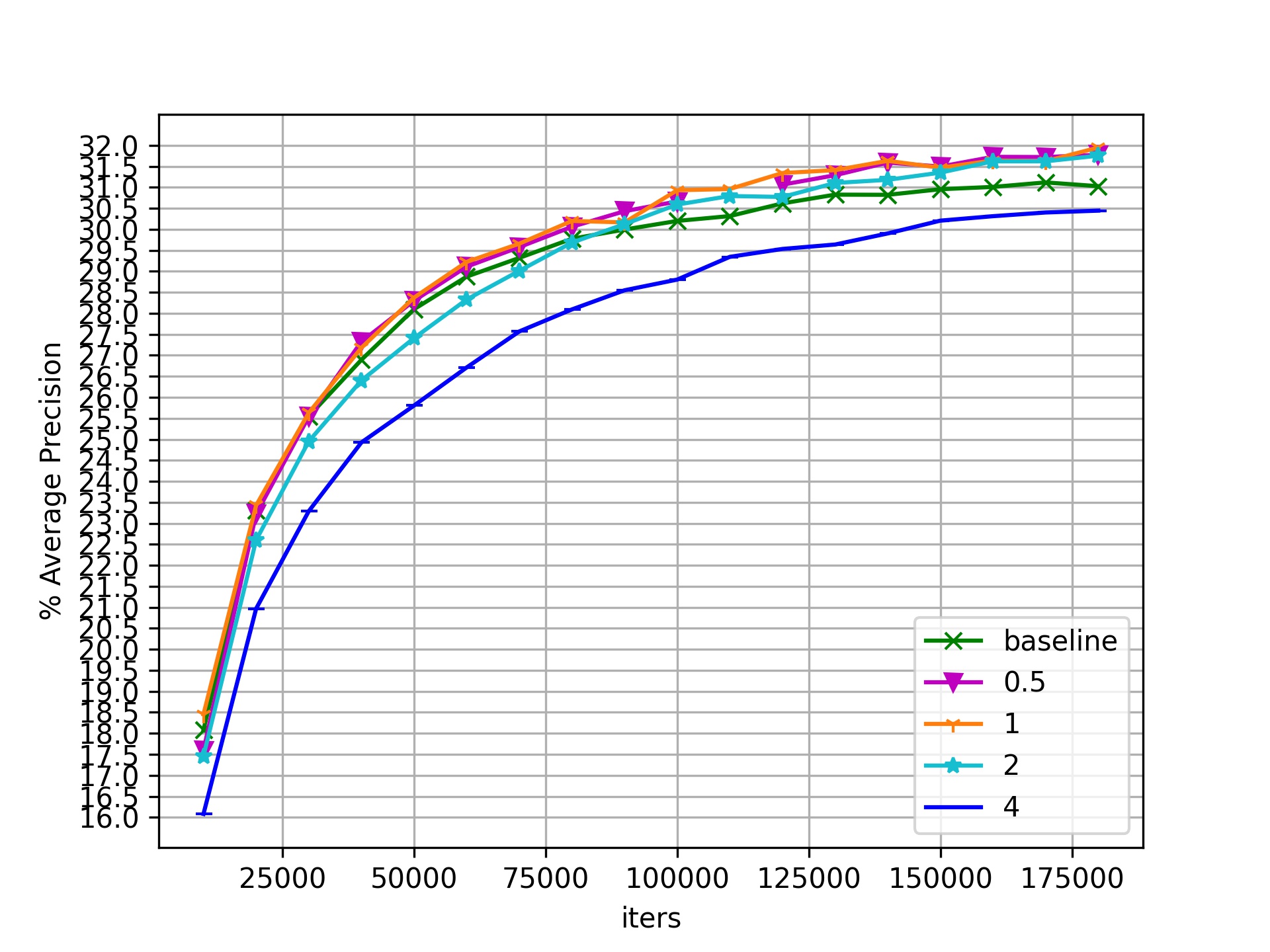}
		\caption{Unsup regression loss weights.}
		\label{fig:ablation_weight_regr_losses}
	\end{subfigure}
	\begin{subfigure}{0.43\linewidth}
		\includegraphics[width=0.9\linewidth]{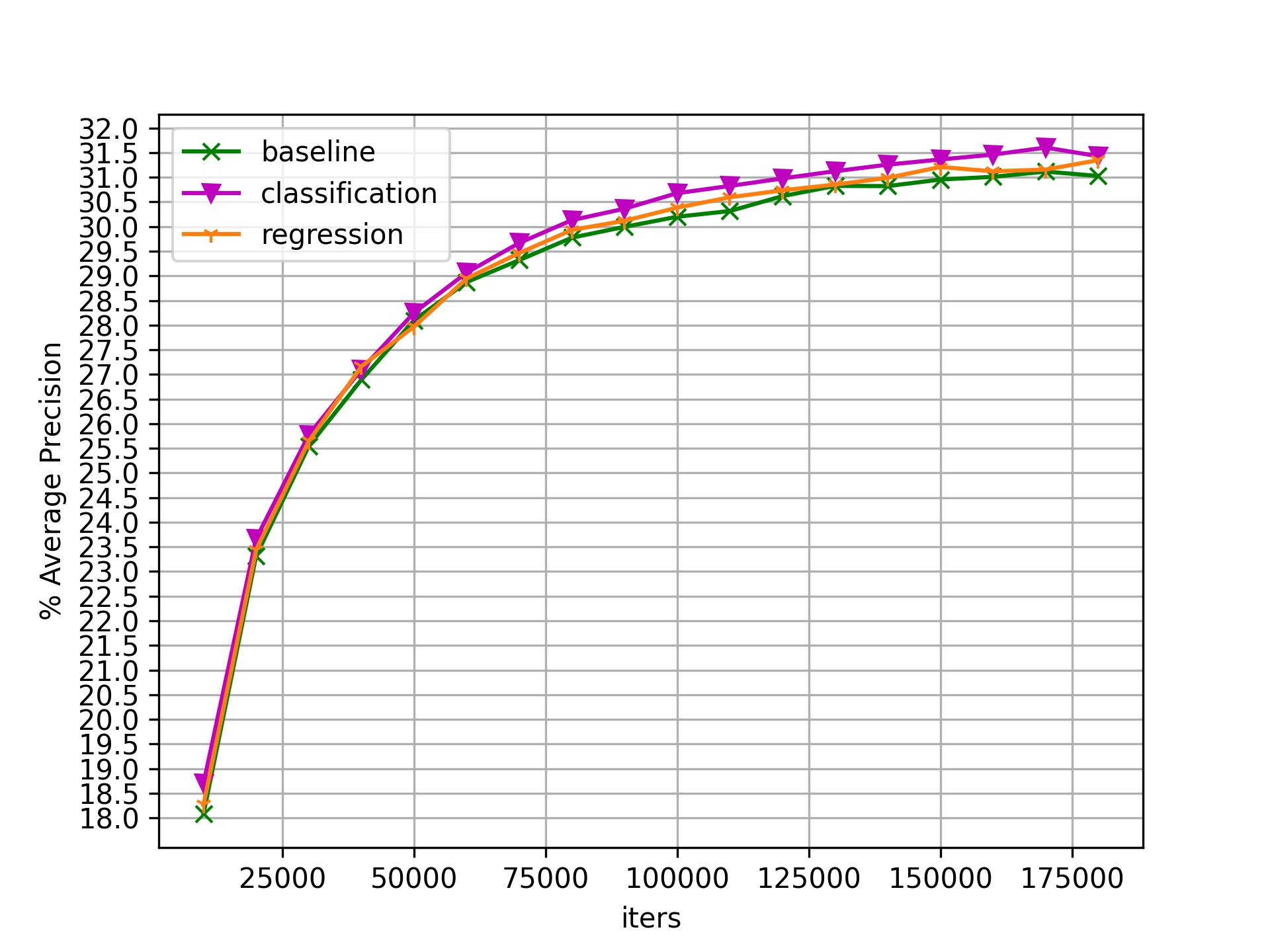}
		\caption{Classification vs Regression.}
		\label{fig:bbox-iou-cls-regr-loss}
	\end{subfigure}
	\begin{subfigure}{0.43\linewidth}
		\includegraphics[width=0.9\linewidth]{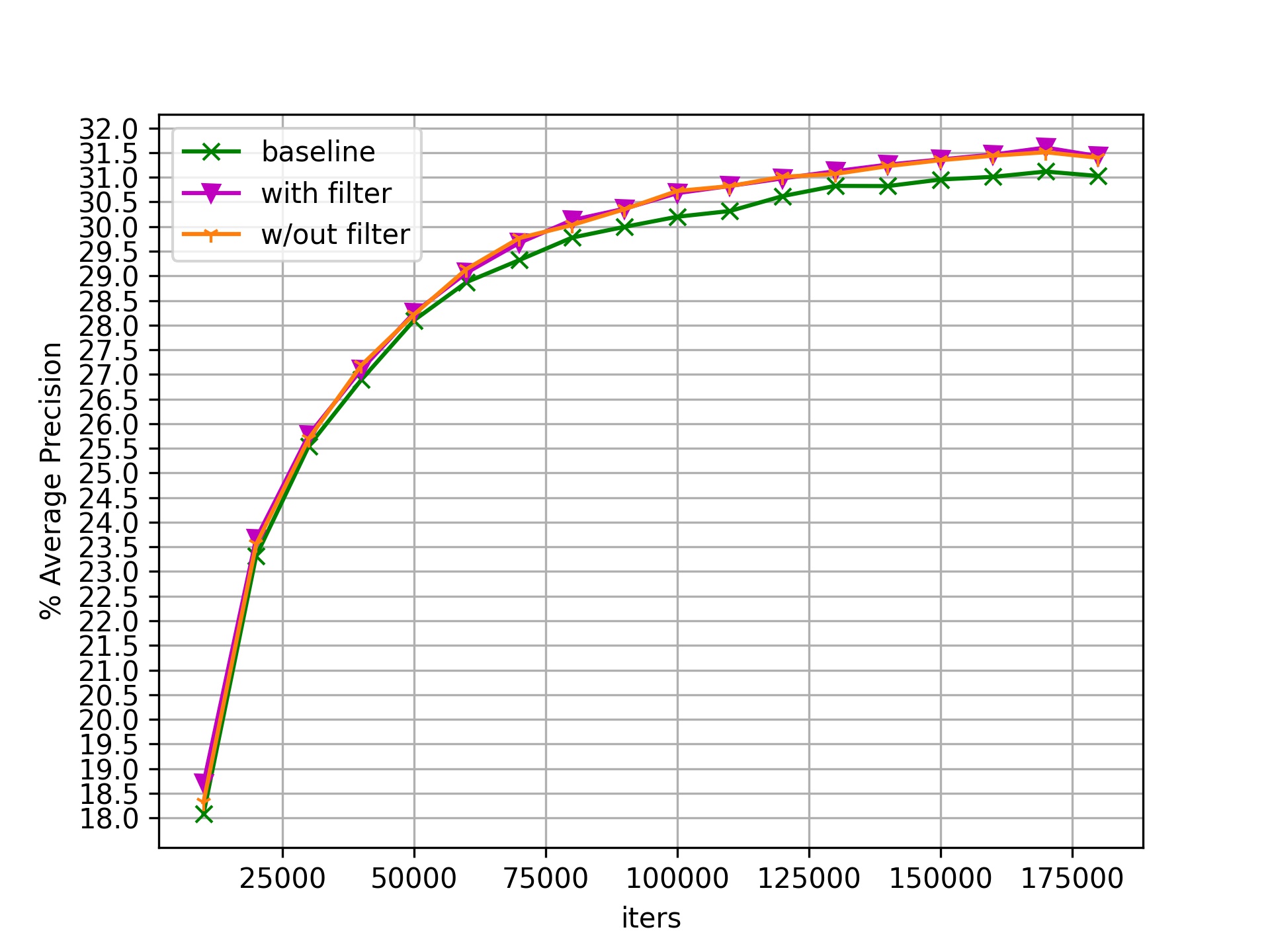}
		\caption{Filtering with bbox IoU score.}
		\label{fig:ablation_with_without_filtering_bbox}
	\end{subfigure}
	\begin{subfigure}{0.43\linewidth}
		\includegraphics[width=0.9\linewidth]{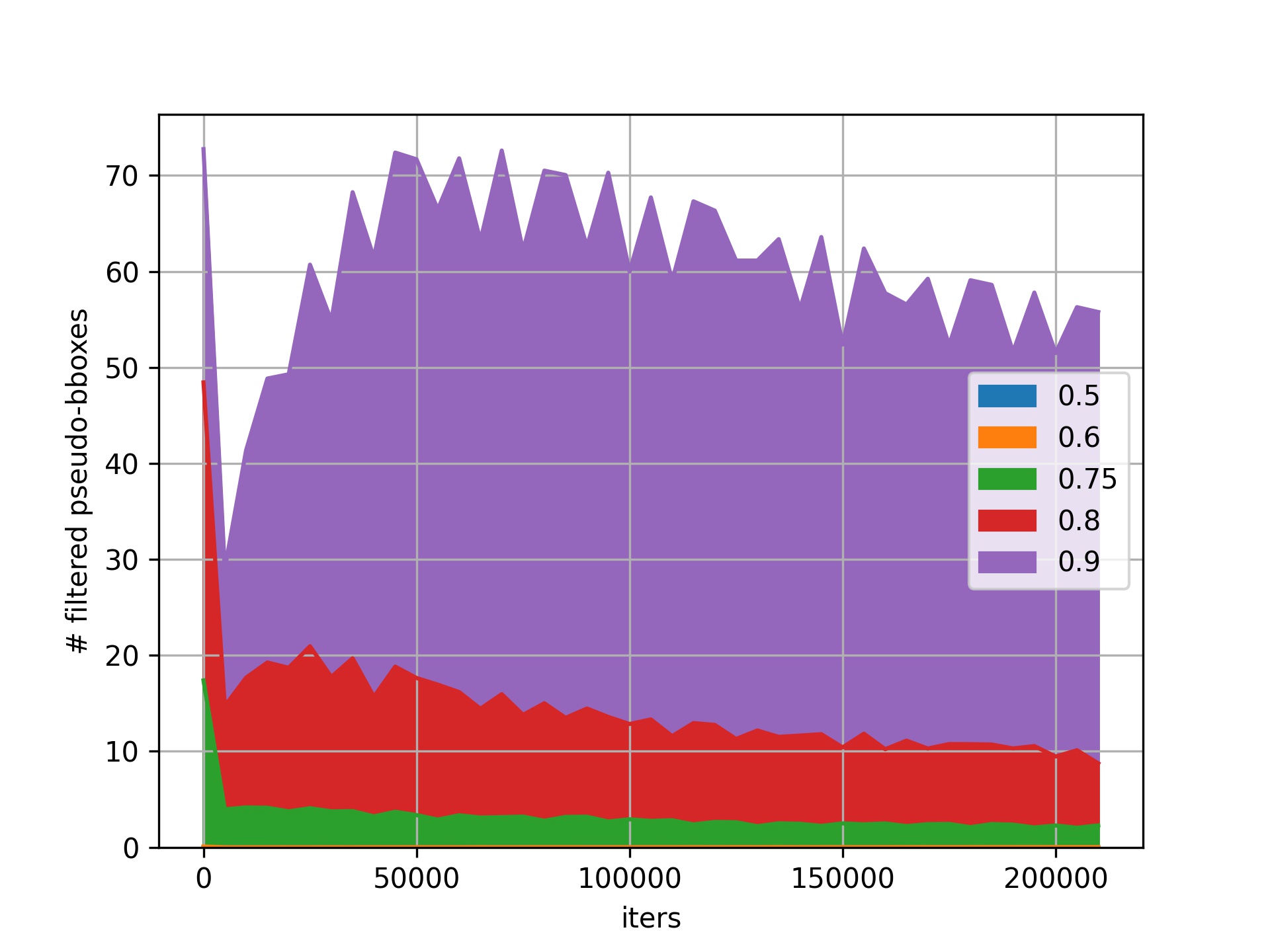}
		\caption{Count filtered pseudo-bboxes.}
		\label{fig:bbox-iou-inf-threshold-count}
	\end{subfigure}
	\caption{Ablation studies:
		(\ref{fig:ablation_weight_regr_losses}) weights for unsupervised regression loss on RPN and RoI.
		(\ref{fig:bbox-iou-cls-regr-loss}) classification vs regression loss on bbox IoU branch.
(\ref{fig:ablation_with_without_filtering_bbox}) filtering bbox on inference with bbox IoU classification score.
		(\ref{fig:bbox-iou-inf-threshold-count}) count bboxes filtered by inference threshold $\mu$ during training.
	}
	\label{fig:ablations}
\end{figure*}

\begin{table*}[bt]
	\footnotesize
	\setlength{\tabcolsep}{1.2pt}
	\begin{center}
		\begin{tabular}{!c|^c|^c|^c|^c|^c|^c|^c|}
			\# & $\beta$ & $AP$ & $AP_{50}$ & $AP_{75}$  & $AP_{s}$ & $AP_{m}$ & $AP_{l}$ \\
			\hline\hline
			0                     & 0.5  & 31.775 & 51.450 & 34.190 & 16.952 & 34.384 & 41.549 \\
			\rowstyle{\bfseries}1 & 1.0  & 31.947 & 51.530 & 34.270 & 16.949 & 34.900 & 41.306 \\
			2                     & 2.0  & 31.754 & 51.078 & 34.143 & 16.691 & 35.008 & 41.670 \\
			%
			3                     & 4.0  & 30.445 & 49.387 & 32.727 & 15.044 & 33.200 & 40.209 \\
		\end{tabular}
	\end{center}
	\caption{Performance varying the weight loss $\beta$ on unsupervised regression losses.}
	\label{tab:ablation_training_regression_unsup}
\end{table*}

\begin{table*}[bt]
	\footnotesize
	\setlength{\tabcolsep}{1.2pt}
	\begin{center}
		\begin{tabular}{c|l|c|c|c|c|c|c|}
			\# & Method & $AP$ & $AP_{50}$ & $AP_{75}$  & $AP_{s}$ & $AP_{m}$ & $AP_{l}$ \\
			\hline\hline
			1 & UT                   & 31.027 & 50.757 & 33.056 & 17.014 & 33.684 & 40.322 \\
			2 & Ours (with filter)    & 31.604 & 51.181 & 33.962 & 16.816 & 34.283 & 40.809 \\
			3 & Ours (w/out filter)   & 31.509 & 51.118 & 33.564 & 16.848 & 34.684 & 40.582 \\
		\end{tabular}
	\end{center}
	\caption{Performance comparison with original Unbiased Teacher (UT) model: (2) Training with BBox IoU branch with and (3) w/out pseudo-labels filtering.}
	\label{tab:ablation_with_without_filtering_bbox}
\end{table*}

\begin{table*}[bt]
	\footnotesize
	\setlength{\tabcolsep}{1.2pt}
	\begin{center}
		\begin{tabular}{!c|^l|^c|^c|^c|^c|^c|^c|}
			\# & $\mu$ & $AP$ & $AP_{50}$ & $AP_{75}$  & $AP_{s}$ & $AP_{m}$ & $AP_{l}$ \\
			\hline\hline
			1 					  & 0.5    & 31.199 & 51.009 & 33.047 & 16.187 & 34.000 & 40.180 \\
			2 					  & 0.6    & 31.128 & 50.785 & 33.268 & 17.102 & 33.805 & 39.932 \\
			3                     & 0.7    & 31.461 & 51.319 & 33.637 & 16.714 & 34.217 & 40.100 \\
			\rowstyle{\bfseries}4 & 0.75   & 31.604 & 51.181 & 33.962 & 16.816 & 34.283 & 40.809 \\
			5 					  & 0.8    & 31.336 & 50.601 & 33.707 & 16.327 & 34.180 & 40.476 \\
			6 					  & 0.9    & 27.125 & 43.515 & 28.800 & 12.815 & 29.486 & 36.034 \\
		\end{tabular}
	\end{center}
	\caption{Performance using BBox IoU classification branch with inference threshold $\theta$ fixed to 0.5 and varying training threshold $\mu$.}
	\label{tab:ablation_training_threshold}
\end{table*}

\begin{table*}[bt]
	\footnotesize
	\setlength{\tabcolsep}{1.2pt}
	\begin{center}
		\begin{tabular}{!c|^c|^c|^c|^c|^c|^c|^c|}
			\# & $\theta$ & $AP$ & $AP_{50}$ & $AP_{75}$  & $AP_{s}$ & $AP_{m}$ & $AP_{l}$ \\
			\hline\hline
			0 					  & 0.3   & 31.404 & 51.205 & 33.792 & 16.273 & 34.542 & 40.851 \\
\rowstyle{\bfseries}1             & 0.4   & 31.630 & 51.185 & 34.044 & 17.387 & 34.494 & 40.784 \\
			2 					  & 0.5   & 31.604 & 51.181 & 33.962 & 16.816 & 34.283 & 40.809 \\
			3					  & 0.6   & 31.158 & 50.227 & 33.418 & 16.203 & 33.687 & 40.323 \\
			4 					  & 0.7   & 30.649 & 49.216 & 33.138 & 16.532 & 33.409 & 39.542 \\
		\end{tabular}
	\end{center}
	\caption{Performance using BBox IoU classification branch with training threshold $\mu$ fixed to 0.75 and varying inference threshold $\theta$.}
	\label{tab:ablation_evaluation_threshold}
\end{table*}


\begin{table*}[bt]
	\footnotesize
	\setlength{\tabcolsep}{1.2pt}
	\begin{center}
		\begin{tabular}{!c|^c|^c|^c|^c|^c|^c|^c|^c|^c|^c|}
			\# & $L_{reg}^{unsup}$ & $x^{sh}$ & scores & deltas  & $AP$ & $AP_{50}$ & $AP_{75}$  & $AP_{s}$ & $AP_{m}$ & $AP_{l}$ \\
			\hline\hline
			1         &            &            &            &            & 31.027 & 50.757 & 33.056 & 17.014 & 33.684 & 40.322 \\
			2         & \checkmark &            &            &            & 31.947 & 51.530 & 34.270 & 16.949 & 34.900 & 41.306 \\
			3         & \checkmark & \checkmark &            &            & 31.754 & 51.189 & 34.032 & 16.850 & 34.657 & 41.320 \\
\rowstyle{\bfseries}4 & \checkmark & \checkmark & \checkmark &            & 32.166 & 51.772 & 34.765 & 16.647 & 34.999 & 41.870 \\
			5         & \checkmark & \checkmark & \checkmark & \checkmark & 31.923 & 51.464 & 34.070 & 16.202 & 35.197 & 41.368 \\
			6         &            & \checkmark & \checkmark & \checkmark & 31.630 & 51.185 & 34.044 & 17.387 & 34.494 & 40.784 \\
		\end{tabular}
	\end{center}
	\caption{Study on unsupervised regression losses and IoU classification loss.}
	\label{tab:net_full}
\end{table*}

\subsection{Ablation study}
\label{ablation-study}

\noindent\textbf{Unsupervised regression loss}.
In this experiment, we empirically show that we can also use regression losses on RPN and RoI head for the unsupervised part.
We test different weights for the constant $\beta$ in the loss formula (see eq. \ref{eq:all_losses}).
In Figure \ref{fig:ablation_weight_regr_losses} and Table \ref{tab:ablation_training_regression_unsup}, we can see that greatly amplifying the contribution can be deleterious, becoming counterproductive in the case of $\beta$ equal to 4.\\

\noindent\textbf{Bounding box IoU branch loss type}.
Our proposal involves a new IoU classification task, trained with a binary cross-entropy function.
In this experiment, we test how performance changes in case our new branch learns a regression task instead of a classification, using a \textit{smooth L1} loss.
In this case, the ground-truth is represented by the real IoU value between the bbox and the ground-truth.
In Figure \ref{fig:bbox-iou-cls-regr-loss}, we can see that classification branch is a bit more stable and reaches a slightly higher performance.\\

\noindent\textbf{With and without filtering bboxes}.
The bbox IoU branch learns to recognize high quality bounding boxes and, as the default behavior, also to pre-filter Teacher's pseudo-bboxes depending on our new threshold score.
In Figure \ref{fig:ablation_with_without_filtering_bbox} and in Table \ref{tab:ablation_with_without_filtering_bbox} (rows \#2 and \#3), we see that our new branch contributes to the increase of the general performance (+0.48\% mAP).
Then, another small improvement is given by the filtering phase, increasing the performance by +0.1\% mAP.\\

\noindent\textbf{Bounding box training threshold $\mu$}.
In this experiment, we test the bbox IoU classification branch, setting inference threshold $\theta$ to 0.5 and varying the training threshold $\mu$.
From Table \ref{tab:ablation_training_threshold}, it is clear that the choice of a correct threshold greatly influences the performance. On the one hand, if the threshold is too low, it does not help the network to learn more descriptive feature maps. On the other hand, if it is too high, the risk to wrongly filter out the bounding boxes will increase.
As we can see in Fig. \ref{fig:bbox-iou-inf-threshold-count}, with the increase of IoU threshold $\mu$, the number of teacher pseudo-bboxes filtered during the training increases exponentially.
This is likely due to an imbalance in training, where the higher the threshold, the fewer high-quality examples are available.
The best value is in the middle between the threshold $u$ (0.5) and the IoU maximum value 1.0.

\noindent\textbf{Bounding box inference filter threshold $\theta$}.
In this experiment, we test our bbox IoU branch, setting the training threshold $\mu$ to 0.75 (best value previously found) and varying the inference threshold $\theta$.
In Table \ref{tab:ablation_evaluation_threshold}, we see that the best value for this threshold is in the middle as expected because the branch is trained to reply 1 if the bbox is good enough and 0 otherwise.\\


\noindent\textbf{IL-net: Improving Localization net}.
Finally, we test the full architecture IL-net, composed of the unsupervised regression losses and the new IoU classification branch.
Since, using both of them, the contribution of the new branch is absorbed by the loss of unsupervised regression (see rows 2 and 5 in Table \ref{tab:net_full}), we performed an ablation study reducing the values in input to the new branch (see eq. \ref{eq:iou_branch}).
This analysis has allowed us to highlight that by removing the contribution of the deltas, we can increase the general performance.
This behavior could be explained by the fact that the deltas are optimized from both losses ($L_{usup}^{reg}$ and $L^{IoU}$), causing the conflict as a result.


%

\section{Conclusions}

In this paper, we proposed two new architectural enhancements with respect to the network proposed in \cite{liu2021unbiased}: a new bounding box IoU classification task to filter out errors on pseudo-labels produced by the Teacher and the introduction of the unsupervised regression losses.
For the former, we introduced a lightweight branch to predict the bounding box IoU quality.
For the latter, we demonstrated how to successfully integrate it in the training, balancing the training tasks.
Our new model called IL-net, which contains both, increases the general SSOD performance by a 1.14\% AP on COCO dataset in a limited annotation regime.

\subsection*{Acknowledgments}
This research benefits from the HPC (High Performance Computing) facility of the University of Parma, Italy.

\bibliographystyle{splncs04}
\bibliography{paper}

\end{document}